\documentclass[runningheads]{llncs}
\usepackage[T1]{fontenc}
\usepackage{graphicx}
\usepackage{siunitx}
\usepackage{placeins}
\usepackage{multirow}
\usepackage{float}
\usepackage{breakurl}
\usepackage{hyperref}
\usepackage{url}
\usepackage{xcolor}
\usepackage{soul}

\title{Voice Pathology Detection Using Phonation}
\author{
    Sri Raksha Siva \inst{1} \and 
    Nived Suthahar \inst{1} \and 
    Prakash Boominathan \inst{2} \and 
    Uma Ranjan \inst{1}
}
\institute{
  Faculty of Engineering and Technology \and
  Faculty of Audiology and Speech Language Pathology \\
   Sri Ramachandra Institute of Higher Education and Research \\
   Chennai, India
}

\begin{document}
\authorrunning{Sri Raksha Siva et al}
\maketitle
%
\begin{abstract}
\sloppy
Voice disorders significantly affect communication and quality of life, requiring an early and accurate diagnosis. Traditional methods like laryngoscopy are invasive, subjective, and often inaccessible. This research proposes a noninvasive, machine learning-based framework for detecting voice pathologies using phonation data.

Phonation data from the Saarbrücken Voice Database are analyzed using acoustic features such as Mel Frequency Cepstral Coefficients (MFCCs), chroma features, and Mel spectrograms. Recurrent Neural Networks (RNNs), including LSTM and attention mechanisms, classify samples into normal and pathological categories. Data augmentation techniques, including pitch shifting and Gaussian noise addition, enhance model generalizability, while preprocessing ensures signal quality. Scale-based features, such as Hölder and Hurst exponents, further capture signal irregularities and long-term dependencies.

The proposed framework offers a noninvasive, automated diagnostic tool for early detection of voice pathologies, supporting AI-driven healthcare, and improving patient outcomes.
\fussy
\end{abstract}


\section{Introduction}
Voice disorders, or pathologies, impair an individual’s ability to produce speech, arising from causes such as vocal cord nodules, infections, neurological impairments, or vocal misuse. These conditions affect communication, psychological well-being, and quality of life, presenting challenges to professionals in vocally demanding roles. Rapid diagnosis and treatment are hindered by the limited availability of efficient and accessible diagnostic tools.

Among voice pathologies, one of the most common and treatable pathologies is that of voice nodules. Voice nodules are most commonly the result of overuse and improper use of the voice. If left untreated, voice nodules can grow in size and eventually run the risk of turning cancerous.

Clinical diagnostic methods include laryngoscopy and stroboscopy, where the vocal cords activity is  visualized.  However, these methods are invasive, equipment-intensive, and dependent on subjective clinical judgment. These limitations underscore the need for noninvasive, objective approaches to enable early detection and intervention.

Machine learning (ML) and artificial intelligence (AI) technologies provide promising solutions by analyzing phonation data, audio recordings of vocalizations, to detect subtle abnormalities. ML-based tools are noninvasive, reduce reliance on specialized equipment, and offer accessibility via platforms like mobile applications.

Common features of characterizing acoustic signals, including voice, include Mel spectrogram, Chroma and MFCC coefficients. In addition, scale based features such as H\"{o}lder and Hurst exponents, have also  been used to analyze signal irregularities and long-term dependencies, adding diagnostic insights. The features used and classification algorithms have a close relationship with each other, and the accuracy of classification depends on a judicious combination of both. There has been limited study on how the same features perform across classifiers, especially time-series based methods such as Recurrent Neural Network (RNN) and its variants. Moreover, the attention mechanism, which has been a cornerstone of many recent developments, has not been investigated in the context of voice pathology. 

 This study aims to study the effect of various algorithms and features on voice nodule classification from phonation data. The contributions of the paper are a comparative study of various features and their efficacy across machine learning algorithms. Acoustic features  Mel Frequency Cepstral Coefficients (MFCC), chroma features, and Mel spectrograms are extracted to capture pathology-specific vocal characteristics. Recurrent Neural Networks (RNNs), including Long Short-Term Memory (LSTM) networks and attention mechanisms, are employed to model sequential audio data. A comparison with SVM, a non-sequential stateless model, is also presented

The dataset used is the Saarbr\"{u}cken Voice Data (SVD), which contains voice recordings of over 2000 individuals, making it one of the largest and widely used datasets for voice pathology.

Section 2 presents related work in this area. Section 3 presents the methodology adopted. Results are presented in Section 4, while Section 5 presents conclusion with discussion on future scope.

\section{Related Work}
Voice analysis in clinical settings are traditionally done using features of jitter,  shimmer and Harmonic-to-noise ratio. Jitter refers to variation in frequency, and shimmer refers to variation in amplitude during vocal cycles. However, these features were not found to be reliable indicators of voice quality \cite{jitter_shimmer_voice_quality}. Recent works have all used the Mel frequency coefficients, or MFCC coefficients for studying voice pathologies. Smith {\it et al}~\cite{ref_article3} used artificial neural networks (ANNs) for diagnosing voice disorders using MFCC coefficients, and reported an accuracy of 75\% on the SVD dataset. Markaki and Stylianou~\cite{ref_article15} utilized modulation spectral features to distinguish pathological voices and reported an accuracy of 94\% on the MEEI dataset. 

Zheng et al.~\cite{ref_article5} introduced a deep learning-based classification framework aimed at improving the accuracy of voice pathology detection. Building on this, Kim and Kim~\cite{ref_article4} incorporated recurrent neural networks (RNNs) in conjunction with empirical mode decomposition (EMD) features to capture the temporal dynamics of speech signals more effectively, yielding enhanced performance in classification tasks.

 Gupta {\it et al}~\cite{ref_article7} developed a hybrid approach that integrates multiple deep learning models to facilitate pathology detection using smartphone-acquired data, offering a portable and accessible diagnostic tool. Alhussein and Muhammad \cite{ref_article8} proposed a parallel deep learning architecture capable of real-time voice monitoring, emphasizing the potential of deep models in smart healthcare environments. Addressing the challenge of data imbalance, a common issue in medical datasets, Lee {\it et al} ~\cite{ref_article1} implemented a model enhanced with Synthetic Minority Oversampling Technique (SMOTE), which  improved classification accuracy on imbalanced datasets. 
 
Spectrogram-based analysis has been another direction of study. Arias-Vergara {\it et al}~\cite{ref_article13} utilized multichannel spectrogram representations to capture diverse acoustic properties of pathological speech, enabling more nuanced speech processing. Similarly, AL-Dhief {\it et al}~\cite{ref_article14} demonstrated that handcrafted features, when used with conventional machine learning classifiers, can still deliver competitive results, underscoring the continued relevance of feature engineering in certain contexts. 

Practical tools have also been an area of focus. Lee {\it et al}~\cite{ref_article6} developed a web-based assessment system, enabling remote screening and evaluation of voice disorders.  Di Cesare {\it et al}~\cite{ref_article11} demonstrated that it is possible to integrate a diagnostic application  into a smartphone-based diagnostic application.
However, these methods are far from ready to be applied in a real world setting. A recent paper by 
Tessler {\it et al}~\cite{ref_article12} presents  a systematic review on the role of deep learning in the diagnosis of vocal cord pathologies, and highlights concerns about bias in data.

The Saarbrücken Voice Database is one of the more complete databases, and  has been pivotal in research. Huckvale and Buciuleac~\cite{ref_article2} examined pathology subsets and audio material impacts, while Lee~\cite{ref_article9} provided benchmarks through evaluations of deep learning methods. Cross-disciplinary advances like those by Torres-Velázquez et al.~\cite{ref_article10} on deep learning in medical imaging and Harar {\it et al}~\cite{ref_article16} on preliminary voice pathology detection underscore the broader potential of these technologies.

Prior work in this area has considered primarily a mixture of pathologies and there is very little work on specific pathologies, especially vocal nodules which are treatable at an early stage. This presents a huge gap in research in methods for early diagnosis of voice nodules from simple methods such as phonation. 

\section{Proposed Model}
This research develops a framework for automated voice pathology detection using machine learning techniques. The primary objective is to analyze phonation data to differentiate between normal and pathological voices, with a focus on the early detection of benign nodules. By employing pre-processing methods, feature extraction techniques, and neural network architectures, the study addresses challenges related to diagnostic accuracy and accessibility.

\subsection{Data Collection}
Phonation data were sourced from the Saarbrücken Voice Database (SVD), a repository widely used in voice pathology research. The database comprises recordings of sustained phonations (e.g., vowel sounds), sentence readings, and conversational speech from individuals with various vocal disorders and healthy controls. Each recording includes metadata such as speaker demographics, diagnostic information, and recording conditions, ensuring diversity and applicability to real-world diagnostic scenarios.

To address class imbalance, data augmentation techniques such as pitch shifting, time stretching, and Gaussian noise addition were applied. 

\subsection{Data Preprocessing}

To ensure quality, consistency  and suitability for machine learning analysis, a number of pre-processing steps were performed. Noise filtering and Silence trimming removed prolonged silent sections with threshold-based algorithms, retaining relevant data, while amplitude normalization addressed variability from recording conditions, ensuring dataset uniformity.

Recordings were segmented into 20–40 millisecond frames to capture short- and long-term temporal dependencies, enhancing the analysis of dynamic vocal properties. Outlier handling corrected or excluded irregular data points to maintain dataset integrity. Additionally, spectral smoothing reduced sharp frequency fluctuations while preserving significant features critical for nodule detection.

These preprocessing steps established a consistent, high-quality dataset for robust feature extraction and voice nodule classification, effectively addressing noise, silence, amplitude variations, segmentation, and outliers.

\begin{figure}[h!]
    \centering
    \includegraphics[width=0.8\textwidth]{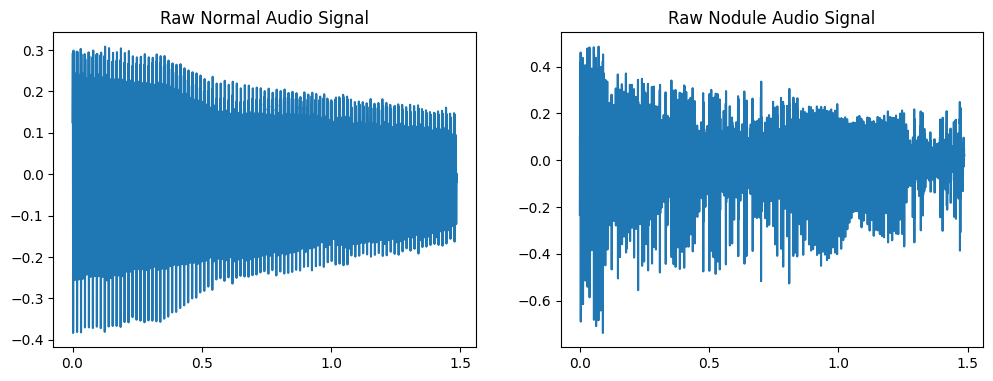}
    \caption{Comparison of Raw Audio Signals: (Left) Normal audio signal showing consistent amplitude variations; (Right) Nodule audio signal exhibiting irregular patterns and higher amplitude fluctuations indicative of vocal fold pathologies.}
    \label{fig:raw_audio_signal}
\end{figure}

\subsection{Feature Extraction}

Feature extraction aimed to capture the acoustic, temporal, and spectral characteristics of phonation data, utilizing a combination of audio features and mathematical techniques for comprehensive signal representation. Mel Frequency Cepstral Coefficients (MFCCs) were employed to encode the timbral and textural properties of speech by emphasizing perceptually relevant frequencies. Derived through the Fast Fourier Transform, a Mel filter bank, and logarithmic spectral energy computation, MFCCs effectively captured subtle vocal anomalies indicative of pathological conditions.

Chroma features, representing the intensity distribution of pitch classes over time, analyzed harmonic and tonal content to detect pitch irregularities linked to voice disorders. Additionally, Mel spectrograms were computed to provide time-frequency visualizations, highlighting energy distributions and noise patterns associated with vocal nodules. To further enhance diagnostic precision, scale-based features were included. Hölder exponents quantified local irregularities, identifying abrupt signal changes, while Hurst exponents measured long-term dependencies and self-similarity, distinguishing between stable and erratic phonation patterns.

The extracted features were systematically evaluated for their relevance in differentiating normal and nodule voices. This feature engineering approach ensured a robust and accurate representation of phonation characteristics, supporting reliable detection of voice nodules.

\subsection{Data Splitting}

The dataset was split into training (70\%), validation (15\%), and testing (15\%) subsets using a stratified approach to preserve class distribution across normal and nodule samples. The training set optimized model parameters, while the validation set monitored performance and guided hyperparameter tuning, helping to mitigate overfitting. The testing set served as an independent benchmark for assessing generalizability and diagnostic accuracy.

Cross-validation further enhanced evaluation robustness by iteratively dividing the training set into folds, ensuring stable and reliable performance metrics. This balanced and systematic approach provided a strong foundation for developing a generalizable voice nodule diagnostic model.
\subsection{Model Creation}

The proposed framework utilized a hybrid architecture combining Recurrent Neural Networks (RNNs) with Long Short-Term Memory (LSTM) units and Convolutional Neural Networks (CNNs) to analyze the temporal and spectral characteristics of phonation data.

\begin{itemize}
    \item \textbf{Recurrent Neural Networks with LSTM}: LSTMs addressed vanishing gradient issues in traditional RNNs by using memory cells and gating mechanisms, enabling the model to retain critical information over extended periods. This was essential for identifying sustained phonation irregularities.

    \item \textbf{Attention Mechanism}: An attention mechanism prioritized key audio segments indicative of nodules while minimizing less relevant data, improving interpretability and diagnostic precision.

    \item \textbf{Convolutional Neural Networks (CNNs)}: CNNs processed spectral features like Mel Spectrograms, capturing spatial dependencies in the frequency domain. This complemented LSTMs by providing insights into the multidimensional nature of phonation data.

    \item \textbf{Hybrid Architecture}: The hybrid model integrated CNNs for spatial feature extraction and LSTMs for temporal pattern modeling. The final layer used a sigmoid activation function to classify samples as normal or nodule, providing probabilistic predictions.
\end{itemize}

Extensive experimentation with architectural parameters, such as the number of layers, kernel sizes, dropout rates, and hidden units, ensured optimal performance. Grid search was used for hyperparameter tuning, optimizing metrics like accuracy, precision, recall, and F1-score. Regularization techniques, including dropout and batch normalization, mitigated overfitting and enhanced generalizability. This hybrid architecture effectively leveraged both temporal and spectral features for reliable voice nodule detection.

\subsection{Model Training}
The training process was designed to enable effective learning from phonation data, with 70\% of the dataset used for training and 15\% reserved for validation to mitigate risks of overfitting and underfitting. The model employed the Adam optimizer, known for its efficiency with sparse gradients and adaptive learning rates, while binary cross-entropy loss measured the difference between predicted probabilities and actual class labels, supporting accurate classification.

To enhance generalizability, dropout layers were implemented to deactivate random neurons during training, reducing overfitting, and batch normalization was applied to standardize input distributions, ensuring faster convergence and stable learning. Hyperparameter tuning, conducted via grid search, optimized key parameters such as learning rate, batch size, and number of epochs. Early stopping terminated training when validation loss plateaued, preventing overfitting and reducing computational overhead.

Training proceeded over multiple epochs, with forward and backward passes updating model parameters to minimize loss. Metrics including accuracy, precision, recall, and F1-score were monitored after each epoch to evaluate performance and inform adjustments. Robustness was ensured through k-fold cross-validation, which divided the data into subsets alternately used for validation and training. Averaged results from all folds provided a comprehensive assessment of model consistency and performance.

The final trained model demonstrated high accuracy and precision on both validation and testing sets, confirming its effectiveness and readiness for application in automated voice nodule detection.

\subsection{Model Validation}

The validation process evaluated the generalization capabilities and reliability of the proposed hybrid model in detecting voice nodules. The validation set, comprising 15\% of the dataset, was reserved exclusively for monitoring the model's performance during training. The class distribution within the validation set was maintained to ensure a balanced and realistic assessment of the model’s ability to distinguish between normal and nodule samples.

To improve robustness, k-fold cross-validation was employed during training. This approach divided the training data into $k$ subsets (folds), with each fold alternately used for validation while the remaining folds were used for training. Performance metrics such as accuracy, precision, recall, and F1-score were averaged across all folds to minimize the influence of any single subset on the overall evaluation.

Overfitting was mitigated through the use of early stopping, which terminated training when validation loss failed to improve after a set number of epochs. Regularization techniques, including dropout layers and batch normalization, were applied to stabilize the training process and enhance the model’s generalizability by reducing noise and addressing potential overfitting.

The validation phase revealed consistent performance metrics, with minimal differences between training and validation results. This consistency underscored the effectiveness of the attention mechanism in highlighting critical temporal features within phonation data. The results demonstrated the model’s robustness and reliability, confirming its readiness for testing on an independent dataset.

\subsection{Hölder and Hurst Exponent Analysis}

The Hölder and Hurst exponents were incorporated to analyze the dynamic properties of phonation signals. These scale-based features provided a quantitative assessment of local irregularities and long-term dependencies in the audio signals, offering insights into nodule vocal characteristics.

\subsubsection{Hölder Exponent}

The H\"{o}lder exponent quantified local irregularities in the phonation signals, identifying abrupt changes or disruptions in vocal patterns often associated with pathological conditions, such as nodules or polyps. The computational approach involved segmenting the audio signals into small temporal intervals and evaluating waveform smoothness within each interval. Normal samples exhibited smoother transitions, reflected by higher H\"{o}lder exponent values, while nodule samples showed abrupt variations, indicated by lower values.

\begin{figure}[ht]
    \centering
    \includegraphics[width=0.6\linewidth]{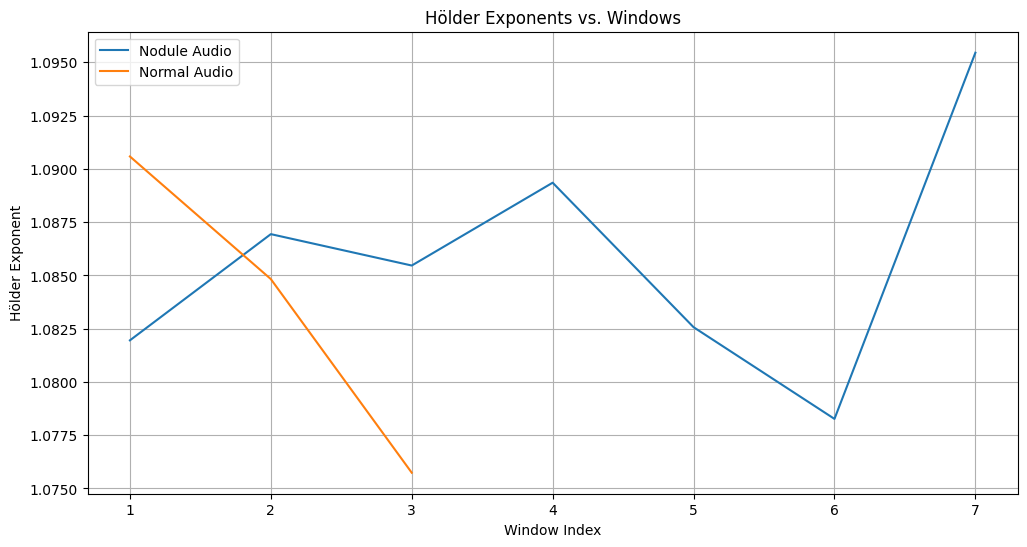}
    \caption{H\"{o}lder Exponent Trends: Variation of Holder exponent values across different audio windows for normal and nodule-affected phonation signals. Normal audio exhibits higher values, reflecting smoother transitions.}
    \label{fig:holder_graph}
\end{figure}

\subsubsection{Hurst Exponent}

The Hurst exponent measured long-term dependencies and self-similarity within phonation signals. It provided a broader view of signal persistence, differentiating between stable and erratic vocal patterns. Normal samples were associated with higher Hurst exponent values, indicating stable and periodic vocal behavior, whereas nodule samples exhibited lower values, reflecting chaotic and irregular phonation patterns. Fractal-based computational techniques were used to evaluate correlation and variability across different time scales in the audio signals.

\begin{figure}[!]
    \centering
    \includegraphics[width=0.8\linewidth]{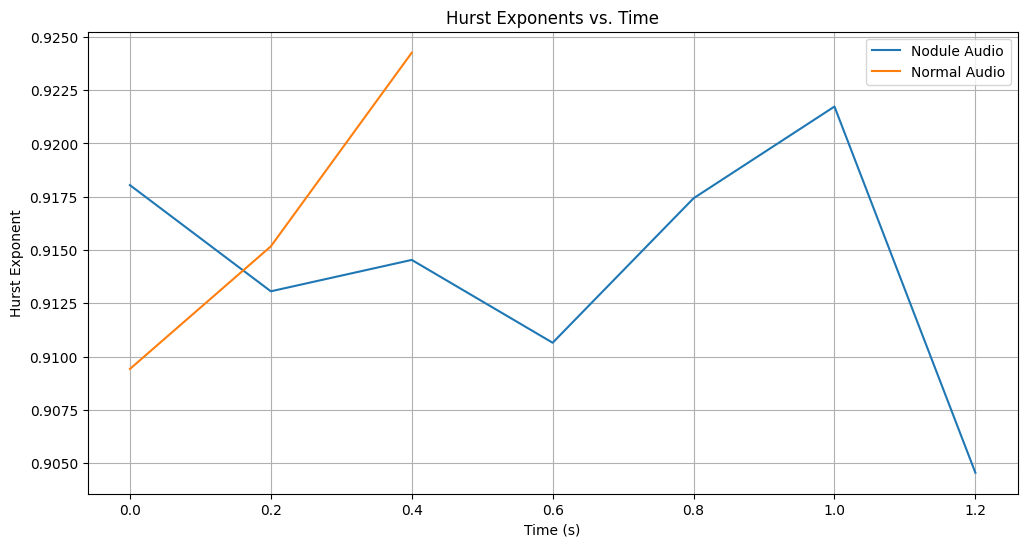}
    \caption{Hurst Exponent Trends: Variation of Hurst exponent values over time for normal and nodule-affected phonation signals. Normal audio demonstrates higher values, indicating stable behavior.}
    \label{fig:hurst_graph}
\end{figure}

\subsubsection{Integration into the Model}

Both exponents were included as features for training the hybrid model. Their addition improved the model’s ability to analyze both micro-level irregularities and macro-level dependencies, enhancing classification performance. Comparative analyses with and without these features highlighted their contribution to achieving higher precision and recall, particularly in cases with subtle nodule indicators.

The integration of H\"{o}lder and Hurst exponents provided a scale-based framework for analyzing complex vocal nodules, supporting their inclusion as key features in non-invasive voice nodule detection.

\subsection{Flow Diagram}
\begin{figure}[h!]
\centering
\includegraphics[width=0.5\textwidth]{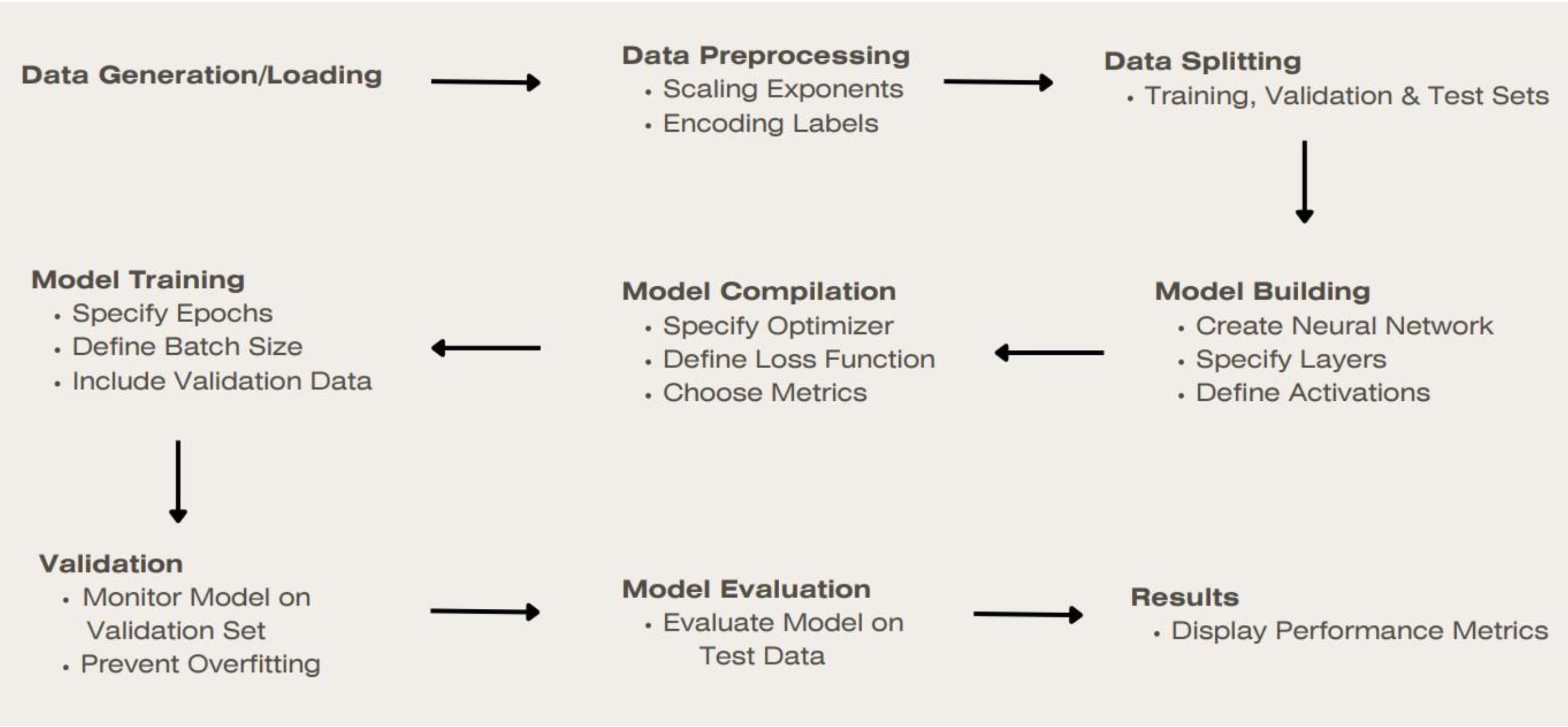}
\caption{Flow Diagram of the Proposed Model}
\label{fig:model_flow}
\end{figure}

The proposed model’s workflow, as illustrated in Figure~\ref{fig:model_flow}, outlines a structured pipeline from data preparation to model evaluation, ensuring a logical progression in the development of the voice nodule detection system. The process begins with loading phonation data from the Saarbr\"ucken Voice Database, enriched with metadata, followed by essential data preprocessing steps such as noise filtering, silence trimming, amplitude normalization, and segmentation into uniform time frames. Additionally, label encoding and scaling of exponents prepare the dataset for feature extraction and analysis.

The dataset is then split into training, validation, and testing subsets, ensuring balanced representation and robust evaluation. The model building phase incorporates a hybrid architecture combining recurrent neural networks (RNNs) with attention mechanisms and convolutional neural networks (CNNs) to effectively analyze both temporal and spectral features. Following model compilation, where the Adam optimizer, binary cross-entropy loss, and evaluation metrics are defined, the model training phase utilizes these configurations over multiple epochs, with regularization techniques like dropout and batch normalization mitigating overfitting. Validation ensures the model’s stability and reliability through metrics such as accuracy, precision, recall, and F1-score, while cross-validation further enhances robustness. Finally, the trained model is evaluated on the test set, and its results are analyzed and reported, confirming its real-world applicability.

\section{Results}
The study demonstrated the effectiveness of various models and feature sets in detecting voice nodules. By leveraging Mel Spectrograms, Chroma Features, MFCCs, and mathematical exponents, the proposed framework provided detailed insights into pathological phonation characteristics. This section presents model performance, spectral differences, and diagnostic implications.

The results for each of the models for the combined features is presented in Table~\ref{table:allacc} 

\begin{table}[ht]
\centering
\fcolorbox{black}{yellow!80}{
    \begin{tabular}{|l|c|c|c|}
    \hline 
    Model & Mean test  Accuracy & Confidence Interval & Standard Deviation \\
    \hline 
        Simple RNN & 0.9266 & (0.9233, 0.9300) & 0.0088 \\
        RNN + Attention & 0.8559 & (0.8524, 0.8594) & 0.0092 \\
        LSTM & 0.9286 & (0.9249, 0.9323) & 0.0098 \\
        LSTM + Attention & 0.7855 & (0.7821, 0.7889) & 0.0089 \\
        SVM & 0.8562 & (0.8523, 0.8600) & 0.0101 \\
        CNN & 0.9313 & (0.9278, 0.9348) & 0.0093 \\
         \hline
         \end{tabular}
         }
         \vspace*{0.1in} 
      \caption{Comparison across models for all features combined}
      \label{table:allacc}
\end{table}

\subsection{Model Performance}
The models combined Mel Spectrograms, Chroma Features, and MFCCs with machine learning frameworks like Simple RNN, RNN with Attention, CNN, and LSTM. Key results include:

\begin{itemize}
    \item \textbf{RNN with Attention}: Paired with Mel Spectrograms, this model achieved the highest performance with a precision of 89.7\%, recall of 85.7\%, and F1-score of 85.1\%, effectively capturing time-frequency dependencies in nodule data.
    \item \textbf{Chroma Features}: Detected tonal inconsistencies, achieving a test accuracy of 78.57\% with RNN + Attention.
    \item \textbf{MFCC Features}: Processed with LSTMs, MFCCs achieved a test accuracy of 71.43\%, demonstrating the importance of capturing long-term vocal dynamics.
\end{itemize}

Mel Spectrograms provided the most comprehensive representation, while Chroma and MFCC features uniquely contributed to detecting tonal and timbral properties. Attention mechanisms enhanced diagnostic precision, reducing false positives and negatives.

\subsection{Spectral Analysis: Male Normal vs. Nodule}
\begin{figure*}[!t]
    \centering
    \includegraphics[width=0.7\textwidth]{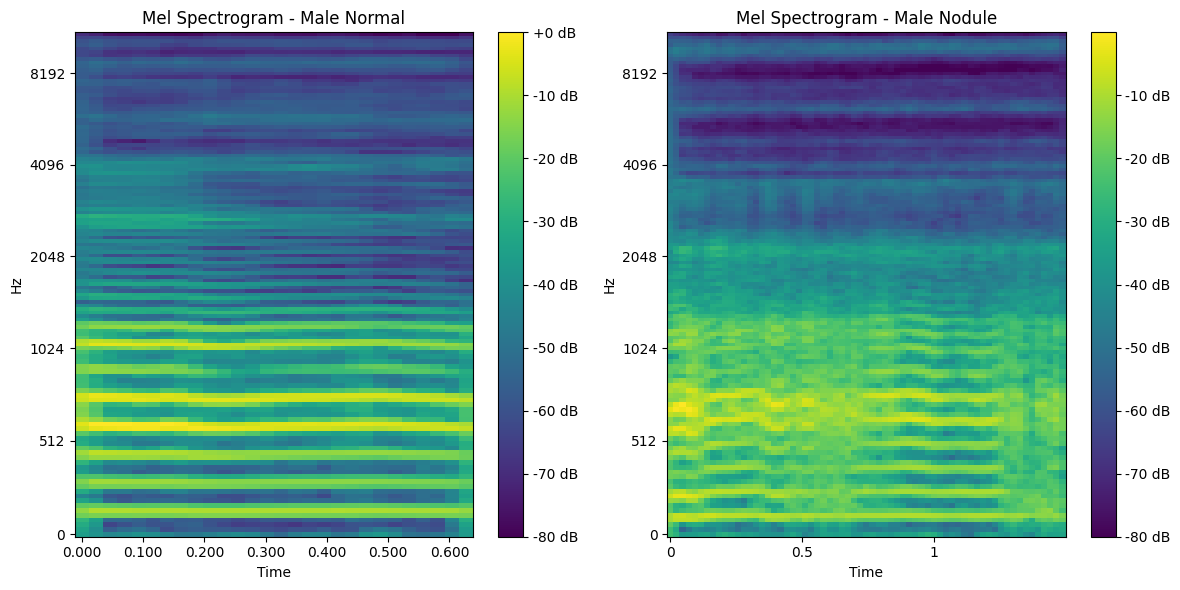}
    \caption{Comparison of Mel Spectrograms for Male Normal (left) and Male Nodule (right) samples. Normal phonation displays smooth and harmonic energy distributions, while nodule-affected samples show energy concentrated in irregular bands, indicative of disrupted vocal fold behavior.}
    \label{fig:mel_spectrogram_male}

    \centering
    \includegraphics[width=0.7\textwidth]{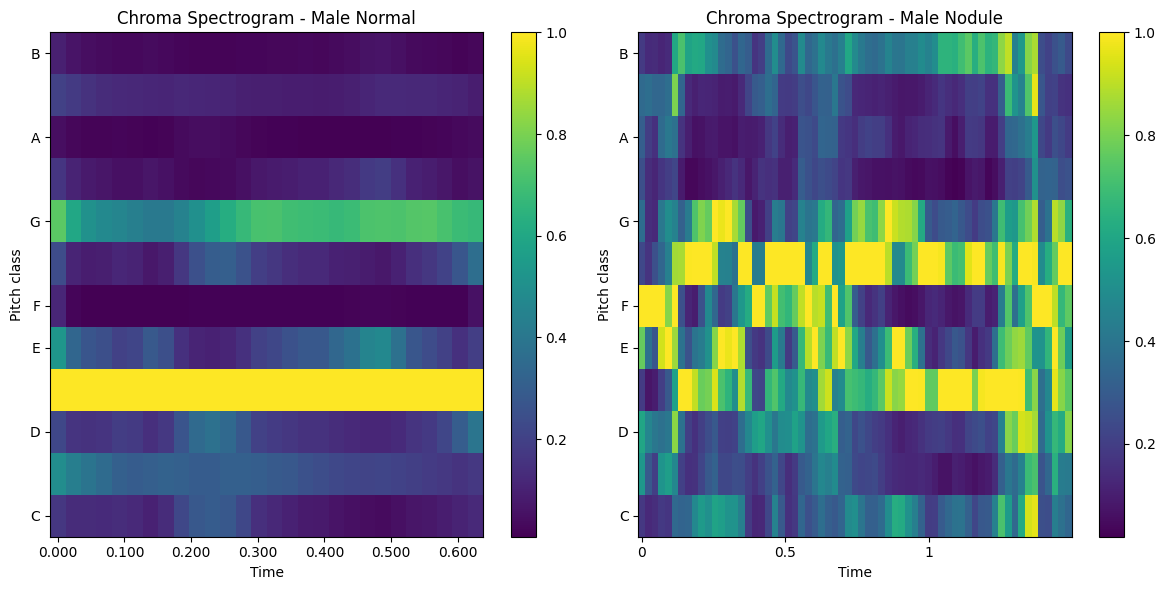}
    \caption{Comparison of Chroma Spectrograms for Male Normal (left) and Male Nodule (right) samples. Normal phonation shows stable pitch class intensities with minimal fluctuations, while nodule-affected samples exhibit erratic pitch transitions, indicating tonal instability.}
    \label{fig:chroma_spectrogram_male}

    \centering
    \includegraphics[width=0.7\textwidth]{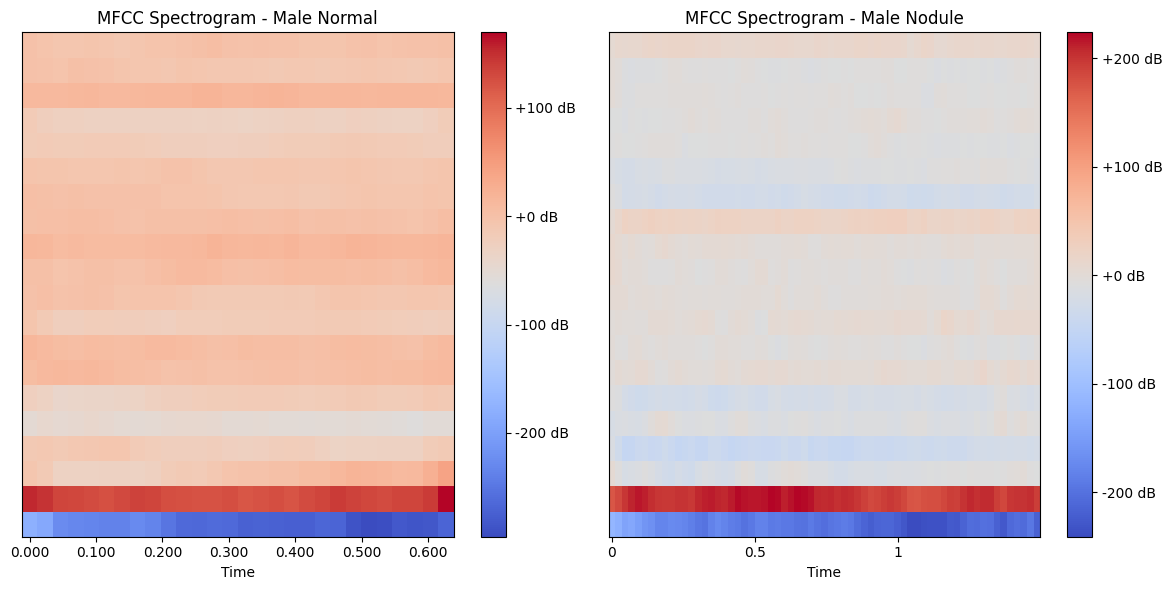}
    \caption{Comparison of MFCC Spectrograms for Male Normal (left) and Male Nodule (right) samples. Normal samples display consistent spectral coefficients with stable timbral properties, whereas nodule samples show fragmented and distorted MFCC patterns, indicative of irregular vocal fold vibrations.}
    \label{fig:mfcc_spectrogram_male}
\end{figure*}

MFCC, Mel Spectrograms and Chroma Features were plotted and Mel and Chroma revealed distinct patterns in male samples, as seen in Fig~\ref{fig:mel_spectrogram_male}, Fig~\ref{fig:chroma_spectrogram_male} and Fig~\ref{fig:mfcc_spectrogram_male}:

\begin{itemize}
    \item \textbf{Normal samples}: Smooth harmonic energy distributions and stable pitch intensities reflected consistent vocal fold vibrations.
    \item \textbf{Nodule-affected samples}: Irregular energy bands and erratic pitch transitions indicated disrupted vocal fold behavior and tonal instability.
\end{itemize}

\subsection{Spectral Analysis: Female Normal vs. Nodule}
\begin{figure*}[!t]
    \centering
    \includegraphics[width=0.7\textwidth]{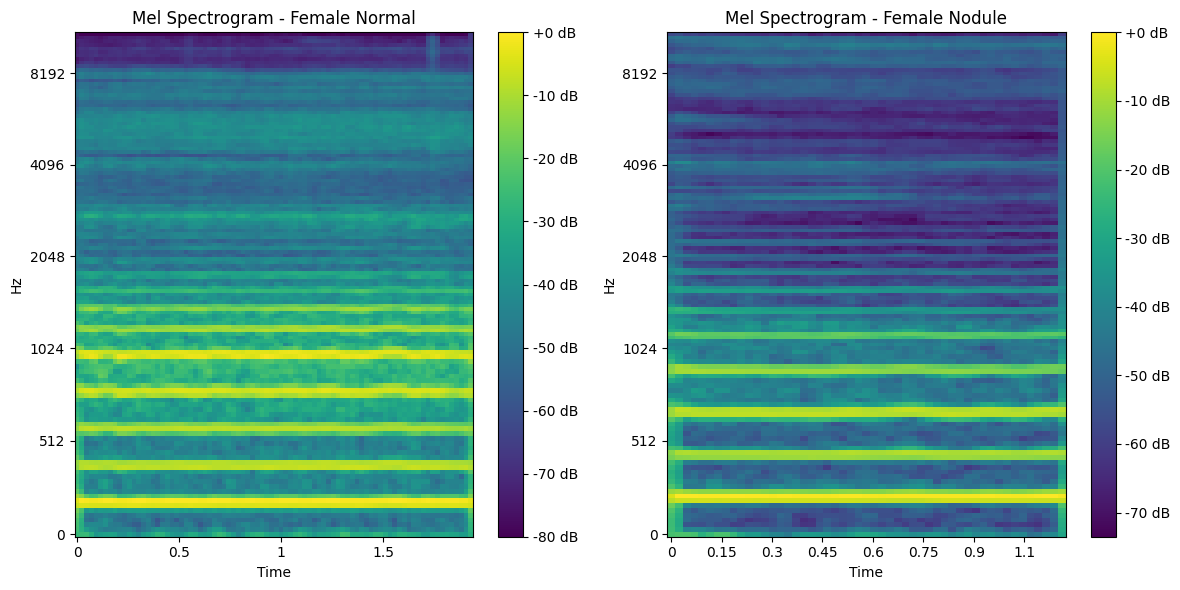}
    \caption{Comparison of Mel Spectrograms for Female Normal (left) and Female Nodule (right) samples. Normal phonation shows smooth and harmonic energy distributions, while nodule-affected samples reveal energy concentrated in irregular bands, indicative of disrupted vocal fold behavior.}
    \label{fig:mel_spectrogram_female}

    \centering
    \includegraphics[width=0.7\textwidth]{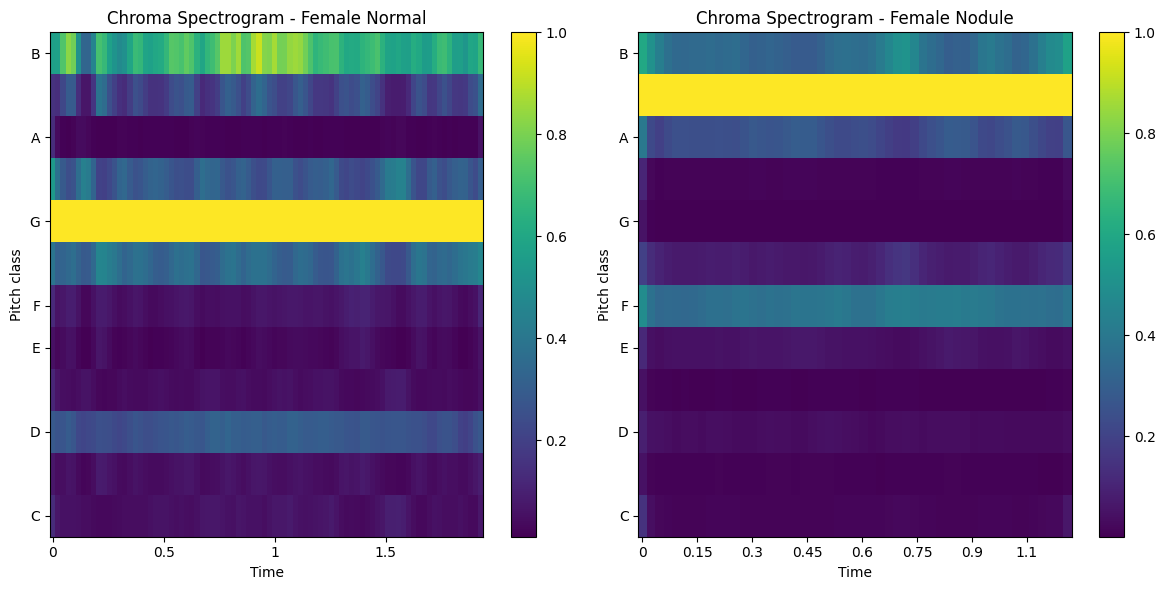}
    \caption{Comparison of Chroma Spectrograms for Female Normal (left) and Female Nodule (right) samples. Normal phonation exhibits stable pitch class intensities, while nodule-affected samples demonstrate erratic pitch transitions, indicating tonal instability.}
    \label{fig:chroma_spectrogram_female}

    \centering
    \includegraphics[width=0.7\textwidth]{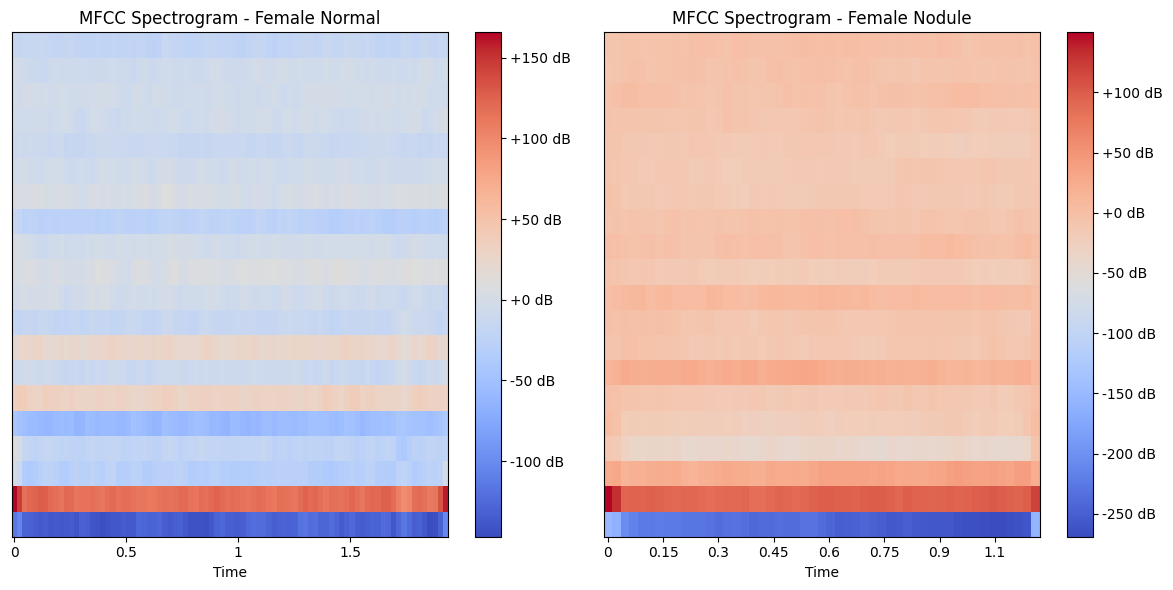}
    \caption{Comparison of MFCC Spectrograms for Female Normal (left) and Female Nodule (right) samples. Normal samples display consistent spectral coefficients with stable timbral properties, whereas nodule samples exhibit fragmented and distorted MFCC patterns, indicative of irregular vocal fold vibrations.}
    \label{fig:mfcc_spectrogram_female}
\end{figure*}

For female samples, MFCCs and Mel Spectrograms identified key differences, as seen from Figures Fig~\ref{fig:mel_spectrogram_female}, Fig~\ref{fig:chroma_spectrogram_female} and Fig~\ref{fig:mfcc_spectrogram_female}:

\begin{itemize}
    \item \textbf{Normal samples}: Stable spectral coefficients and smooth energy distributions highlighted clear vocal properties.
    \item \textbf{Pathological samples}: Fragmented MFCC patterns and disjointed energy indicated irregular vocal vibrations.
\end{itemize}

Chroma Features provided additional insights:
\begin{itemize}
    \item \textbf{Nodule samples}: Sharp pitch fluctuations in nodule samples contrasted with smoother transitions in normal voices, amplified by higher pitch ranges.
\end{itemize}

These analyses confirmed the diagnostic utility of Mel Spectrograms, Chroma Features, and MFCCs across male and female samples.

\subsection{Hölder and Hurst Exponent Analysis}
The inclusion of Hölder and Hurst exponents provided quantitative insights into signal irregularities and dependencies:

\begin{itemize}
    \item \textbf{Hölder Exponent}:
    \begin{itemize}
        \item Normal voices had higher values (e.g., 1.0863), indicating smooth transitions.
        \item Pathological voices showed lower values (e.g., 1.0844), reflecting abrupt disruptions.
    \end{itemize}
    \item \textbf{Hurst Exponent}:
    \begin{itemize}
        \item Normal voices exhibited values around 0.9137, signifying stability.
        \item Pathological voices had slightly higher values (e.g., 0.9156), indicating erratic patterns.
    \end{itemize}
\end{itemize}

\begin{table}[htbp]
\centering
\caption{Model Performance for Different Feature Sets}
\begin{tabular}{|l|l|l|l|}
\hline
\textbf{Feature Set}      & \textbf{Model}          & \textbf{Test Loss} & \textbf{Test Accuracy (\%)} \\ \hline
\multirow{3}{*}{Mel}      & Simple RNN             & 0.2148             & 92.98                       \\ \cline{2-4} 
                          & RNN + Attention        & 0.2104             & 85.71                       \\ \cline{2-4} 
                          & CNN                    & 0.4585             & 85.71                       \\ \hline
\multirow{4}{*}{Chroma}   & Simple RNN             & 0.9548             & 64.29                       \\ \cline{2-4} 
                          & RNN + Attention        & 0.9321             & 78.57                       \\ \cline{2-4} 
                          & LSTM                   & 0.9356             & 78.57                       \\ \cline{2-4} 
                          & LSTM + Attention       & 0.9820             & 78.57                       \\ \hline
\multirow{2}{*}{MFCC}     & Simple RNN             & 0.9548             & 64.29                       \\ \cline{2-4} 
                          & LSTM                   & 0.6792             & 71.43                       \\ \hline
\end{tabular}
\label{tab:model_performance}
\end{table}

\vspace{-1em}
\begin{table}[htbp]
\centering
\caption{Hurst and Hölder Exponents Analysis}
\begin{tabular}{|l|l|l|}
\hline
\textbf{Measure}      & \textbf{Normal} & \textbf{Pathological (Nodule)} \\ \hline
Hurst Exponent        & 0.9137          & 0.9156                         \\ \hline
Hölder Exponent       & 1.0863          & 1.0844                         \\ \hline
\end{tabular}
\label{tab:hurst_holder_analysis}
\end{table}

The pair-wise statistical significance of the tests was performed, and the results are presented in Table~\ref{tab:model_comparison_pvalue}
\begin{table}[h]
    \centering
    \renewcommand{\arraystretch}{1.2}
    \caption{Comparison across models}
    \fcolorbox{black}{yellow!80}{
    \begin{tabular}{|l|c|c|c|}
        \hline
        \textbf{Model Comparison} & \textbf{t-stat} & \textbf{p-value} & \textbf{Cohen's d effect size} \\
        \hline
        Simple RNN vs RNN + Attention & 31.51 & 0.0000 & 7.8 \\
        Simple RNN vs LSTM & -0.82 & 0.4202 & -0.2 \\
        Simple RNN vs LSTM + Attention & 52.85 & 0.0000 & 15.9 \\
        Simple RNN vs SVM & 27.25 & 0.0000 & 7.4 \\
        Simple RNN vs CNN & -2.20 & 0.0357 & -0.5 \\
        \hline
        RNN + Attention vs LSTM & -29.76 & 0.0000 & -7.7 \\
        RNN + Attention vs LSTM + Attn & 32.44 & 0.0000 & 7.8 \\
        RNN + Attention vs SVM & -0.11 & 0.9139 & -0.03 \\
        RNN + Attention vs CNN & -37.16 & 0.0000 & -8.2 \\
        \hline
        LSTM vs LSTM + Attention & 54.53 & 0.0000 & 15.3\\
        LSTM vs SVM   & 30.85  & 0.0000 & 7.3 \\
        LSTM  vs CNN & -1.03 & 0.9139 & -0.28 \\
        SVM vs CNN & -26.40 & 0.0000 & -7.75 \\ 
    \end{tabular}
    }
    \label{tab:model_comparison_pvalue}
\end{table}

\FloatBarrier 

\begin{figure*}[!t]
    \centering
    \includegraphics[width=0.7\textwidth]{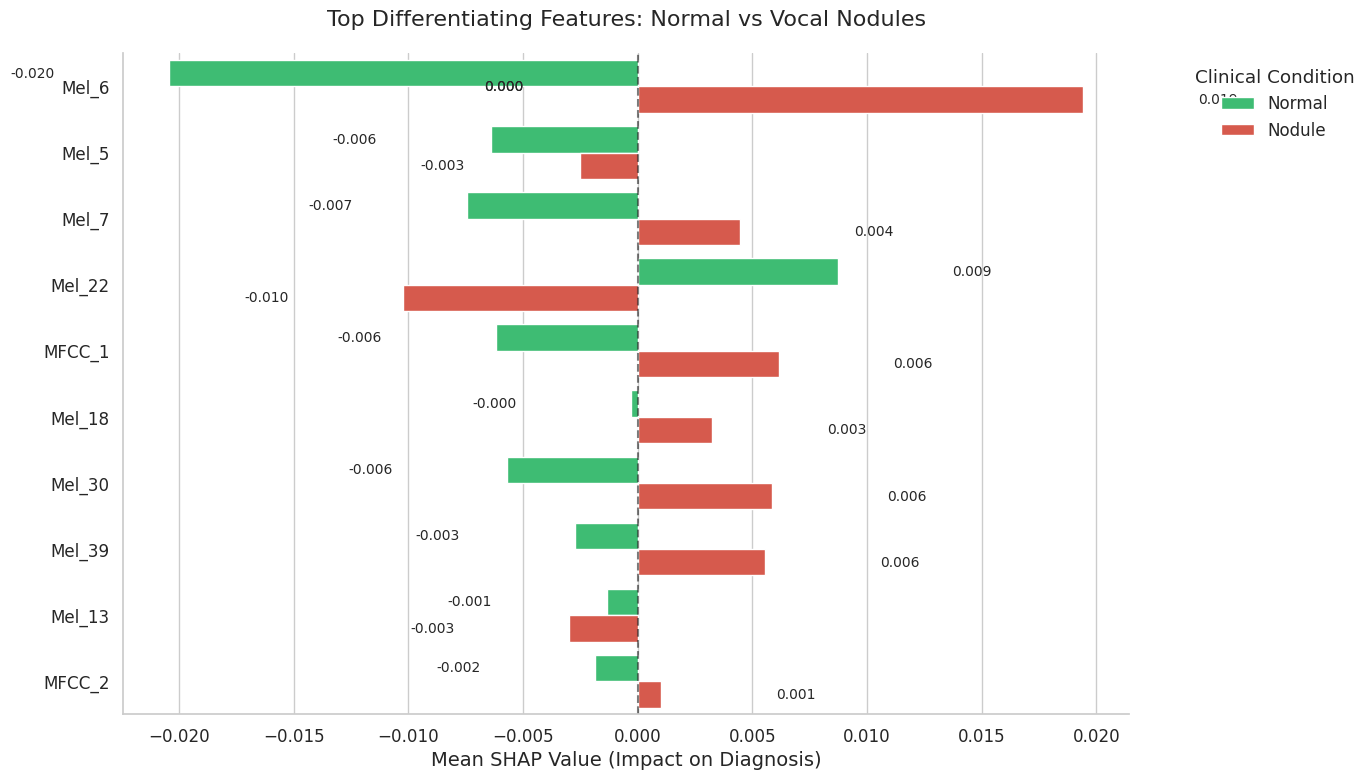}
    \caption{SHAP explainability features}
    \label{fig:shap}
\end{figure*}

To check the explainability of the various features, a SHAP plot was used (Figure~\ref{fig:shap}. It is seen that, as expected, Mel coefficients and specifically the 6th Mel coefficient holds the highest discriminatory power. From these, one can conclude that there is a variation of discriminatory power even within the coefficients, and a more fine-grained approach of picking and choosing coefficients across MEL and MFCC coefficients can give better results than using one or the other. 

\subsection{Key Findings}

The key findings from the study are summarized as follows:

\begin{itemize}
    \item \textbf{Feature Effectiveness}: Mel Spectrograms proved to be the most informative for capturing spectral and temporal details, while Chroma and MFCC features highlighted tonal and timbral properties critical for identifying pitch and harmonic irregularities.
    \item \textbf{Model Architecture}: The attention mechanism in RNNs improved the model's ability to focus on critical temporal regions, enhancing recall and precision. LSTMs effectively captured long-term dependencies, making them suitable for identifying persistent nodule trends.
    \item \textbf{Scale-Based Insights}: The integration of Hölder and Hurst exponents added a valuable layer of diagnostic depth, improving the model’s precision in detecting vocal anomalies.
     \item \textbf{Explainability}: specific coefficients with Mel spectrogram or MFCC hold high discriminatory power. Hence, instead of choosing MFCC or MEL coefficients as a single entity for feature extraction, a further refinement of features is necessary.
\end{itemize}

\subsection{Implications}

The findings of this study have several implications for voice nodule detection:

\begin{itemize}
    \item \textbf{Early Diagnosis}: The models demonstrated the ability to detect subtle pathological changes, offering potential for early intervention and improved patient outcomes.
    \item \textbf{Scalability and Accessibility}: By integrating spectral and scale-based features with machine learning frameworks, the study highlights the feasibility of developing non-invasive diagnostic tools that can be deployed in clinical and remote environments.
    \item \textbf{Foundation for Future Research}: The inclusion of Hölder and Hurst exponents paves the way for further exploration into scale-based and computational approaches to vocal health diagnostics, encouraging interdisciplinary research.
\end{itemize}

This study presents a comprehensive framework for voice nodule detection, bridging the gap between computational advancements and clinical applications. By leveraging advanced spectral features and scale-based methodologies, it contributes to the development of accurate and scalable solutions for non-invasive voice nodule diagnosis.

\section{Conclusion}

This paper studies the effect of various machine learning models and features. Both frequency based features and scale based features have been included in the study. In addition, a study of the relative performance of the models for each feature has been shown. Stability and robustness has also been studied using statistical measures. It is seen that attention mechanism does not work well, while the simple RNN model does. This is possibly due to the non-stationarity of the voice signal, especially in pathological conditions. Attention mechanism assumes a predictive relationship from a previous segment of the signal, and this does not hold in the case of non-stationary processes. The work has been conducted with data from the Saarbrücken Voice Database (SVD). This method, while performing reasonably well, has been tested only on data from German population. Hence, its generalization to other populations has not been studied. This is due to the lack of available data from other populations. Future extensions of this work include cross-validation on diverse populations, and a sub-refinement of the MEL and MFCC coefficients and their validation.

\begin{credits}
\subsubsection{\ackname}  This study was funded by IIT Madras Pravartak PRAYAS grant

\subsubsection{\discintname}
The authors have no competing interests to declare that are
relevant to the content of this article. 
\end{credits}


\begin{thebibliography}{16}
\sloppy 

\bibitem{jitter_shimmer_voice_quality}
Kreiman, Jody and Gerratt, Bruce and Gabelman, Brian: Jitter, shimmer, and noise in pathological voice quality perception, The Journal of the Acoustical Society of America, 112, 2446 (2002)

\bibitem{ref_article3}
Smith, S. S., Abel, J. H., Eibling, D. E.: Artificial neural networks: An emerging method for diagnosis of voice disorders. Journal of Voice , 20 (2), 280--290 (2006).

\bibitem{ref_article15}
Markaki, M., Stylianou, Y.: Voice Pathology Detection and Discrimination Based on Modulation Spectral Features. IEEE Transactions on Audio, Speech, and Language Processing 19 (7), 1938--1948 (2011).

\bibitem{ref_article5}
Zheng, Y., Guo, L., Chen, J., Tan, T.: A deep learning approach to classification of voice disorders. IEEE Journal of Biomedical and Health Informatics 21(1), 7--17 (2017).


\bibitem{ref_article4}
Kim, K. J., Kim, H. Y.: Recurrent neural networks for voice pathology classification using empirical mode decomposition-based features. Expert Systems with Applications 41(7), 3501--3507 (2014).
\bibitem{ref_article7}
Gupta, V., Le, T., Valente, T. W.: An ensemble deep learning approach for automatic voice pathology detection using smartphones. IEEE Journal of Biomedical and Health Informatics 24(2), 512--522 (2020).
\bibitem{ref_article8} M. Alhussein and G. Muhammad, "Automatic Voice Pathology Monitoring Using Parallel Deep Models for Smart Healthcare," in IEEE Access, vol. 7, pp. 46474-46479 (2019)

\bibitem{ref_article1}
Lee, J.-N., Lee, J.-Y.: An Efficient SMOTE-Based Deep Learning Model for Voice Pathology Detection. Applied Sciences, 13 (6), 3571 (2023).
\bibitem{ref_article13}
Arias-Vergara, T., Klumpp, P., Vasquez-Correa, J. C., Nöth, E., Orozco-Arroyave, J. R., Schuster, M.: Multi-Channel Spectrograms for Speech Processing Applications Using Deep Learning Methods. Springer (2020)
\bibitem{ref_article14}
AL-Dhief, F. T., et al.: Voice Pathology Detection Using Machine Learning Technique. IEEE ISTT, pp. 99--104 (2020).
\bibitem{ref_article6}
Lee, C. M., Choi, J. W., Kim, J. S.: Development of a web-based system for voice pathology self-assessment using artificial intelligence. Journal of Voice 33(2), 181--187 (2019).
\bibitem{ref_article11}
Di Cesare, M.G.; Perpetuini, D.; Cardone, D.; Merla, A. Assessment of Voice Disorders Using Machine Learning and Vocal Analysis of Voice Samples Recorded through Smartphones. BioMedInformatics, 4, 549-565 (2024)
\bibitem{ref_article12}
Tessler I, Primov-Fever A, Soffer S, et al. Deep learning in voice analysis for diagnosing vocal cord pathologies: a systematic review. Eur Arch Otorhinolaryngol. 281(2):863-871 (2024)
\bibitem{ref_article2} 
Huckvale, M,  Buciuleac, C.: Automated Detection of Voice Disorder in the Saarbrücken Voice Database. Effects of Pathology Subset and Audio Materials. Proceedings of the Annual Conference of the International Speech Communication Association (2021)
\bibitem{ref_article9}
Lee, J.-Y. Experimental Evaluation of Deep Learning Methods for an Intelligent Pathological Voice Detection System Using the Saarbruecken Voice Database. Appl. Sci. (2021)
\bibitem{ref_article10}
Torres-Velázquez M, Chen WJ, Li X, McMillan AB. Application and Construction of Deep Learning Networks in Medical Imaging. IEEE Trans Radiat Plasma Med Sci. 5(2):137-159.(2021)
\bibitem{ref_article16}
P. Harar, J. B. Alonso-Hernandezy, J. Mekyska, Z. Galaz, R. Burget and Z. Smekal, "Voice Pathology Detection Using Deep Learning: a Preliminary Study," 2017 International Conference and Workshop on Bioinspired Intelligence (IWOBI), Funchal, Portugal, 1-4 (2017)

\fussy 
\end{thebibliography}
\end{document}